\documentclass[10pt,twocolumn,letterpaper]{article}
\usepackage[pagenumbers]{cvpr} 
\usepackage{comment}
\usepackage{amsmath}
\usepackage{multicol}
\usepackage{multirow}
\usepackage{caption}
\usepackage{graphicx}
\usepackage{subcaption}
\def\ie{{\em i.e.}}
\def\eg{{\em e.g.}}

\usepackage[dvipsnames]{xcolor}

\definecolor{cvprblue}{rgb}{0.21,0.49,0.74}
\usepackage[pagebackref,breaklinks,colorlinks,citecolor=cvprblue]
{hyperref}

\def\paperID{6} 
\def\confName{CVPR}
\def\confYear{2024}

\title{Can ChatGPT Detect DeepFakes? A Study of Using Multimodal Large Language Models for Media Forensics}

\author{Shan Jia$^{1}$, Reilin Lyu$^{2}$, Kangran Zhao$^{3}$, Yize Chen$^{3}$, \\Zhiyuan Yan$^{3}$, Yan Ju$^{1}$, Chuanbo Hu$^{4}$, Xin Li$^{4}$, Baoyuan Wu$^{3}$, Siwei Lyu$^{1}$\\
$^1$ University at Buffalo, State University of New York, Buffalo, USA\\
$^2$ Williamsville East High School, Buffalo, USA \\
$^3$ The Chinese University of Hong Kong, Shenzhen, China \\
$^4$ University at Albany, State University of New York, Albany, USA\vspace{-.5em}\\
}

\begin{document}
\maketitle
\begin{abstract}

DeepFakes, which refer to AI-generated media content, have become an increasing concern due to their use as a means for disinformation. Detecting DeepFakes is currently solved with programmed machine learning algorithms. In this work, we \textbf{investigate} the capabilities of multimodal large language models (LLMs) in DeepFake detection. We conducted qualitative and quantitative experiments to demonstrate multimodal LLMs and show that they can expose AI-generated images through careful experimental design and prompt engineering. This is interesting, considering that LLMs are not inherently tailored for media forensic tasks, and the process does not require programming. We discuss the limitations of multimodal LLMs for these tasks and suggest possible improvements.

\end{abstract}    
\section{Introduction}
\label{sec:intro}

The meteoric rise of Generative AI (GenAI) models is one of the most exciting developments in recent years. State-of-the-art GenAI models have demonstrated incredible abilities to create realistic images, audio, and videos\footnote{\eg, Midjourney \url{www.midjourney.com} and Stable Diffusion \url{https://stability.ai/} for image generation, Elevenlab \url{https://elevenlabs.io/} for audio generation, and Pika {\url{https://pika.art/}} and OpenAI's Sora {\url{https://openai.com/sora}} for video generation.} from text prompts. While AI-generated content has numerous beneficial uses, such as in the movie and advertising industry, its misuse to produce deleterious content, commonly known as {\em DeepFakes}, seriously undermines the credibility of information and trust in digital media. As a result, identifying DeepFakes has become a crucial and timely task in media forensics.

The current DeepFake detection is solved by dedicated machine learning algorithms written as coded programs. Most of the existing methods are based on data-driven deep neural network models that are trained on labeled datasets of real and DeepFake media (\eg, Celeb-DF \cite{li_etal_cvpr20}). Detection often relies on statistical features of the media signal, and users must use them through dedicated programming languages, tools, or services. 

Meanwhile, large language models (LLMs) and the conversational agents built upon them, such as ChatGPT, Google Gemini, and the open-source LLaMA, 
have emerged in recent years as versatile tools with wide-ranging applications. The LLM chatbots' intuitive natural language interface significantly eases user interactions and obviates the reliance on programming expertise. More importantly, LLMs have exhibited a strong ability to encode vast knowledge bases from the existing text corpus. This ability has been further extended to images and videos, as the most recent LLMs bring in vision-language models to possess multimodal understanding, as showcased in the most recent ChatGPT based on the GPT4V multimodal LLM \cite{achiam2023gpt}. As such, multimodal LLM chatbots offer a more intuitive and user-friendly means to solve complex problems and have found applications in computer and network forensics \cite{scanlon2023chatgpt}, face verification \cite{deandrestame2024good}, and medical diagnosis \cite{wu2023can}. In a recent unpublished study \cite{shi2024shield}, the authors tested LLMs in identifying facial spoofing and forgery. However, this study focuses on qualitative studies based on a set of individual queries, hence only providing a partial glimpse of the full potential of the LLMs in detecting DeepFakes.
%
\begin{figure}[t]
\centering
\includegraphics[width=1.0\linewidth]{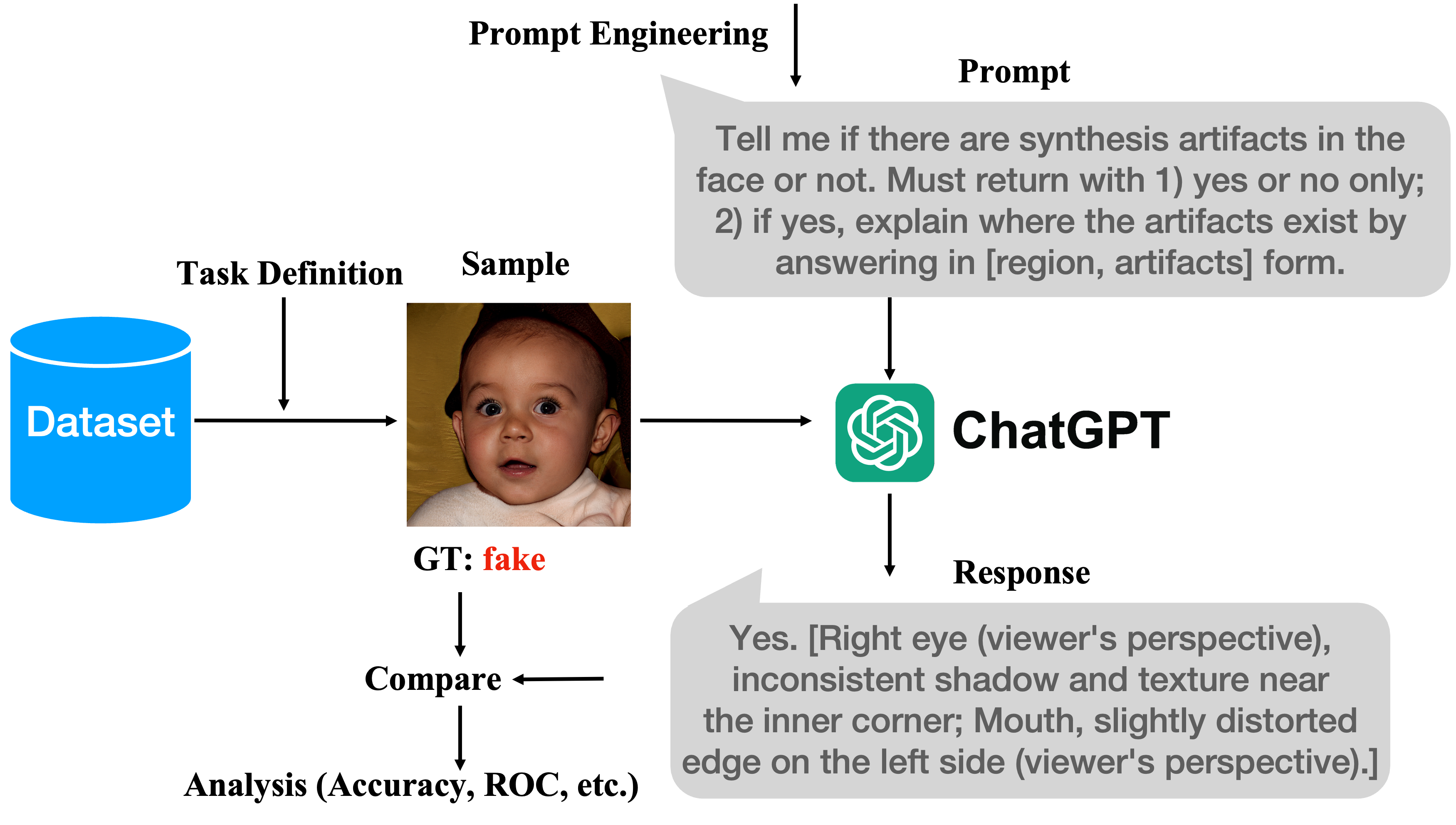}
\caption{\em The overall process of using multimodal LLMs to detect AI-generated face images.}
\label{fig:overview}
\vspace{-0.35cm}
\end{figure}

In this work, our aim is to provide a more comprehensive and quantitative evaluation of the ability of multimodal LLMs to detect DeepFakes. The overall process is illustrated in Fig. \ref{fig:overview}. Specifically, our objective is to demonstrate the feasibility and performance of multimodal LLMs in exposing AI-generated face images. For an input face image, we accompany it with a text prompt that requests a Yes/No response on whether the accompanying image is AI-generated, along with explanations and justifications for the answer. The text prompt is crucial, as it forms the sole interface between the user and the multimodal LLM chatbot for media forensic tasks. Our study focuses on the forms of text prompts that can effectively elicit meaningful responses from LLMs \footnote{All text prompts and results used in this study will be available from \url{https://github.com/shanface33/GPT4MF_UB}.}. On a set of face images, we conduct extensive qualitative and quantitative evaluations of the performance of popular multimodal LLMs on this task. Our initial experiments have yielded several key insights:
\begin{itemize}
    \item Multimodal LLMs demonstrate a certain capability to distinguish between authentic and AI-generated imagery, drawing on their semantic understanding. This discernment is interpretable by humans, offering a more intuitive and user-friendly option compared to traditional machine learning (ML) detection methods.
    \item The efficacy of multimodal LLMs in identifying AI-generated images is satisfactory, with an Area Under the Curve (AUC) score of approximately 75\%. However, their accuracy in recognizing genuine images is noticeably lower. This discrepancy arises because a lack of semantic inconsistencies does not automatically confirm an image's authenticity from the LLMs' standpoint.
    \item The semantic detection capabilities of these LLMs cannot be fully harnessed through simple binary prompts, which can lead to their refusal to provide clear answers. Effective prompting techniques are crucial for maximizing the potential of multimodal LLMs in differentiating between real and AI-generated images.
    \item Presently, multimodal LLMs do not incorporate signal cues or data-driven approaches for this task. While their independence from signal cues enables them to identify AI-created images regardless of the generation model used, their performance still falls short of the latest detection methodologies.
\end{itemize}
We hope that this study will encourage future exploration of the use and improvement of LLMs for media forensics and DeepFake detection. The remainder of the paper is organized as follows. Section \ref{sec:bkg} provides an overview of the relevant literature on LLMs and Deepfake face detection. Section \ref{sec:method} presents the methodology of our study. Comprehensive evaluation results and analysis are given in Section \ref{sec:exp}, and Section \ref{sec:con} concludes the article.. 

\section{Background}
\label{sec:bkg}

\subsection{Large Language Models}
Large Language Models (LLMs) are large-scale foundational deep neural network models (characterized by billions of parameters) that perform natural language-related tasks. Their basic function is to predict the next words in sentences based on previous words. LLMs typically adopt the {\em transformer} architecture \cite{vaswani2017attention}, distinguished by its attention mechanism that evaluates the importance of different words for understanding the text. This architecture provides a more advanced memory structure for handling long-term dependencies than traditional recurrent neural networks, especially when the model was pre-trained on a large text corpus and later fine-tuned with minimal modifications for specific datasets. LLMs are typically trained on gigantic volumes of unlabeled text from the Internet. The training process for LLMs capitalizes on the statistical patterns of human languages and can be subsequently tuned to other applications. 

The popularity of LLMs is largely attributed to the family of {\em generative pretrained transformers} (GPTs) developed by OpenAI. The GPT-1 model, which debuted in 2018, has $117$ million parameters and is the first practical LLM that achieved human-level language understanding in tasks such as textual entailment and reading comprehension. Subsequently, GPT models have quickly evolved with scaled-up capacity and improved performance of task-agnostic and few-shot learning challenges. The GPT4V model, introduced by OpenAI in 2022, has a whopping $175$ billion parameters. Considering that the total corpus on the Internet up to 2022, which more or less represents all human-generated texts throughout history, is about $500$ billion tokens, one can think of the GPT-4 model as a compression model of all human knowledge captured in written texts \cite{deandrestame2024good}. In this sense, it is perhaps not so surprising that GPT -4 can achieve human-level performance in text-understanding tasks. LLM models have recently been extended for cross-modal understanding. In late 2023, OpenAI released the latest GPT-for-vision (GPT4v) model \cite{achiam2023gpt}, which accepts images as input and text prompts. This has been followed up by other LLMs from major companies, such as Google Bard + Gemini \cite{team2023gemini}. 

Ordinary users were exposed to the power of LLMs through conversational agents (chatbots) that use LLMs to engage in natural dialogues for question answering, text summarization, recommendations, and assistance of writing and debugging code, etc. The most well-known LLM-based chatbot is OpenAI's ChatGPT. Since its introduction in November 2022, ChatGPT has rapidly become the fastest growing consumer app ever, having over $100$ million monthly active users within just two months of its release. Besides providing an intuitive conversational user interface, the chatbots also help improve the underlying LLMs by using reinforcement learning from human feedback to gain user feedback.

\subsection{DeepFake Faces: Generation and Detection}

\begin{figure}[t]
    \centering
    \includegraphics[width=0.48\textwidth]{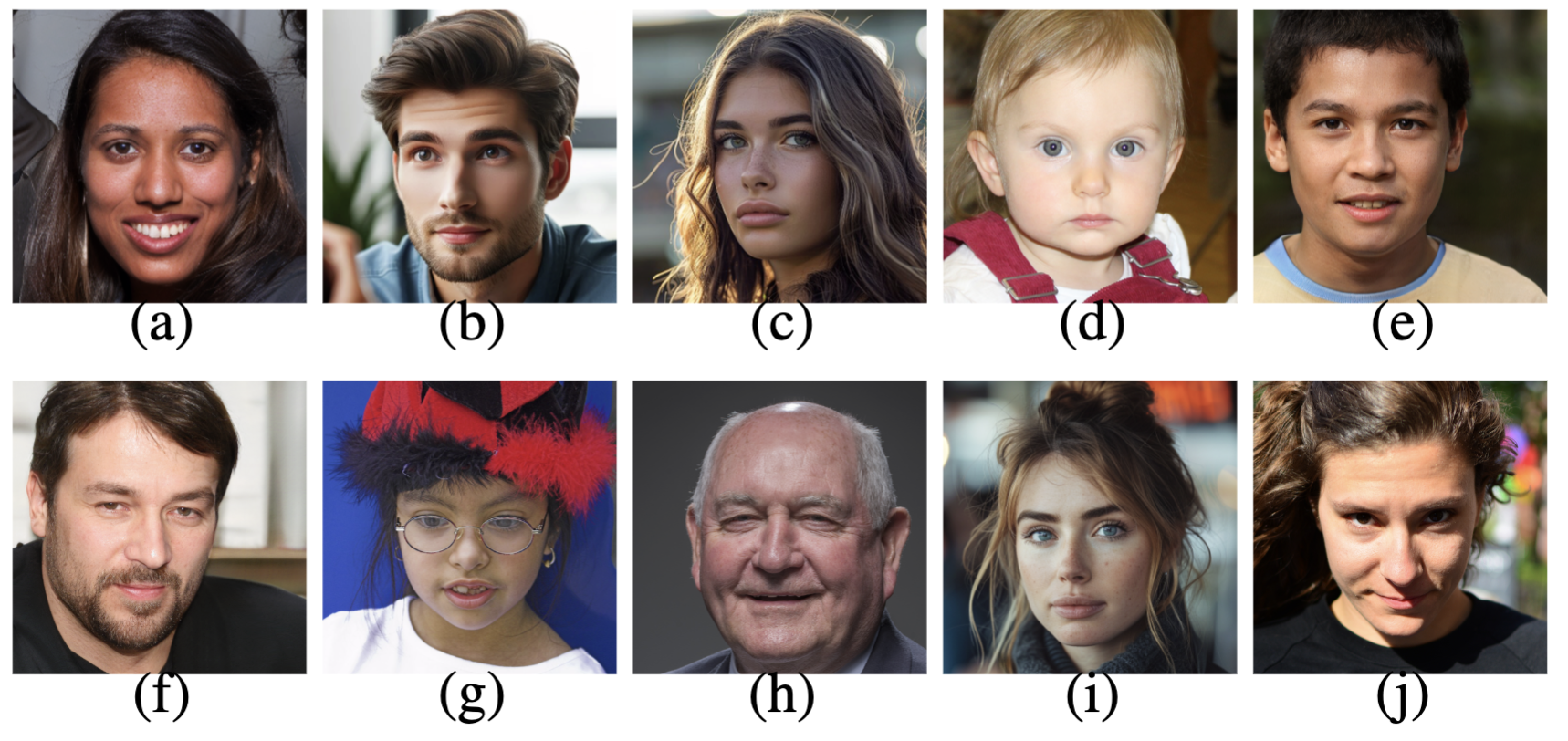}
    \vspace{-0.76cm}
    \caption{\emph{Which of these images are real and AI-generated? \\Answer:}} \vspace{-0.6cm} \rotatebox{180}{\small .(e) Fake, (d) Real, (c) Fake, (b) Fake, (a) Fake} \rotatebox{180}{\small .(j) Real, (i) Fake, (h) Real, (g) Real, (f) Fake}
\vspace{-0.41cm}
    \label{fig:gan-example}
\end{figure}

AI-generated realistic human face images are the earliest and the most well-known examples of DeepFakes. DeepFake faces are created with generative adversarial networks (GANs) and diffusion models. They have a high level of realism in fine details of skin and facial hairs and challenge human's ability to distinguish from images of real human faces (Fig. \ref{fig:gan-example}). DeepFake faces have been used as profile images for fake social media accounts in disinformation campaigns \cite{theverge,cnn1,cnn2,reuters}. 

Existing DeepFake face detection methods are mostly formulated as binary classification problems. Based on the features used, these methods fall into three major categories. Methods in the first category (\eg, \cite{li_etal_wifs18,yang_etal_ih19a,matern2019exploiting,ciftci2020hearts,hu_li_icassp21}) are based on inconsistencies exhibited in the {\em physical/physiological} aspects in the DeepFake images. Methods in the second category (\eg, \cite{mccloskey2018detecting,li2019exposing, li2020face, wolter2022wavelet, yang_ijcai21}) use {\em signal-level} artifacts introduced during the synthesis process. The majority of current detection methods 
(\eg, \cite{nguyen2019capsule, cao2022end, dong2022explaining, ju2022fusing, xu2024mdtl, han2023fcd, corvi2023detection}) fall into {\em data-driven} methods that directly use various DNNs trained on real and DeepFake samples to capture specific artifacts. There also exist several large-scale benchmark datasets to evaluate DeepFake detection performance \cite{roessler2019faceforensics++, wang2019cnn-generated, ju2023glff, corvi2023detection}. 

Current DeepFake face detection methods are typically developed using programming languages like Python and specialized libraries to construct neural network models or other machine learning algorithms (\eg, Scikit-Learn, PyTorch, TensorFlow). These models are then trained on datasets of labeled data. However, the programming language interface represents a significant hurdle for both the developers and users of these detection algorithms.

\section{Methodology}
\label{sec:method}

\definecolor{color1}{rgb}{0.31, 0.43, 0.48}
\definecolor{color2}{rgb}{0.945,  0.816, 0.804}
\definecolor{color3}{rgb}{0.851, 0.906, 0.839}

Our study aims to evaluate the utility and efficacy of multimodal LLMs in media forensics, and we choose the problem of identifying AI-generated images of human faces as the main focus. The rationale is as follows. Firstly, while the multimodal LLMs are technically equipped to analyze video and audio content, their optimal performance is observed with images. Secondly, detecting realistic DeepFake face images is one of the most thoroughly studied topics. It can be used to compare the capabilities of a multimodal LLM with state-of-the-art methods. Thirdly, prior research identified a wealth of semantic indicators. Human can identify semantic inconsistencies in faces, making the study much more accessible to viewers. We can use these established semantic cues to craft targeted prompts to enhance detection efficacy. We choose to use OpenAI's GPT4V Vision model (\ie, GPT4V-vision-preview \footnote{\url{https://platform.openai.com/docs/guides/vision}}) as the subject of the study. 
It provides an API that greatly streamlines experimental procedures, especially for Python-based implementations. This feature is instrumental in simulating conversational contexts on a large scale. We design experiments in which GPT4V model assesses whether a face image is AI-generated based on the text prompts in Fig. \ref{fig:overview}. We also consider Google Gemini 1.0 Pro API for comparison (note that the Gemini web app has restrictions on analyzing images containing human faces).

\begin{figure}[t]
    \centering
    \includegraphics[width=0.48\textwidth]{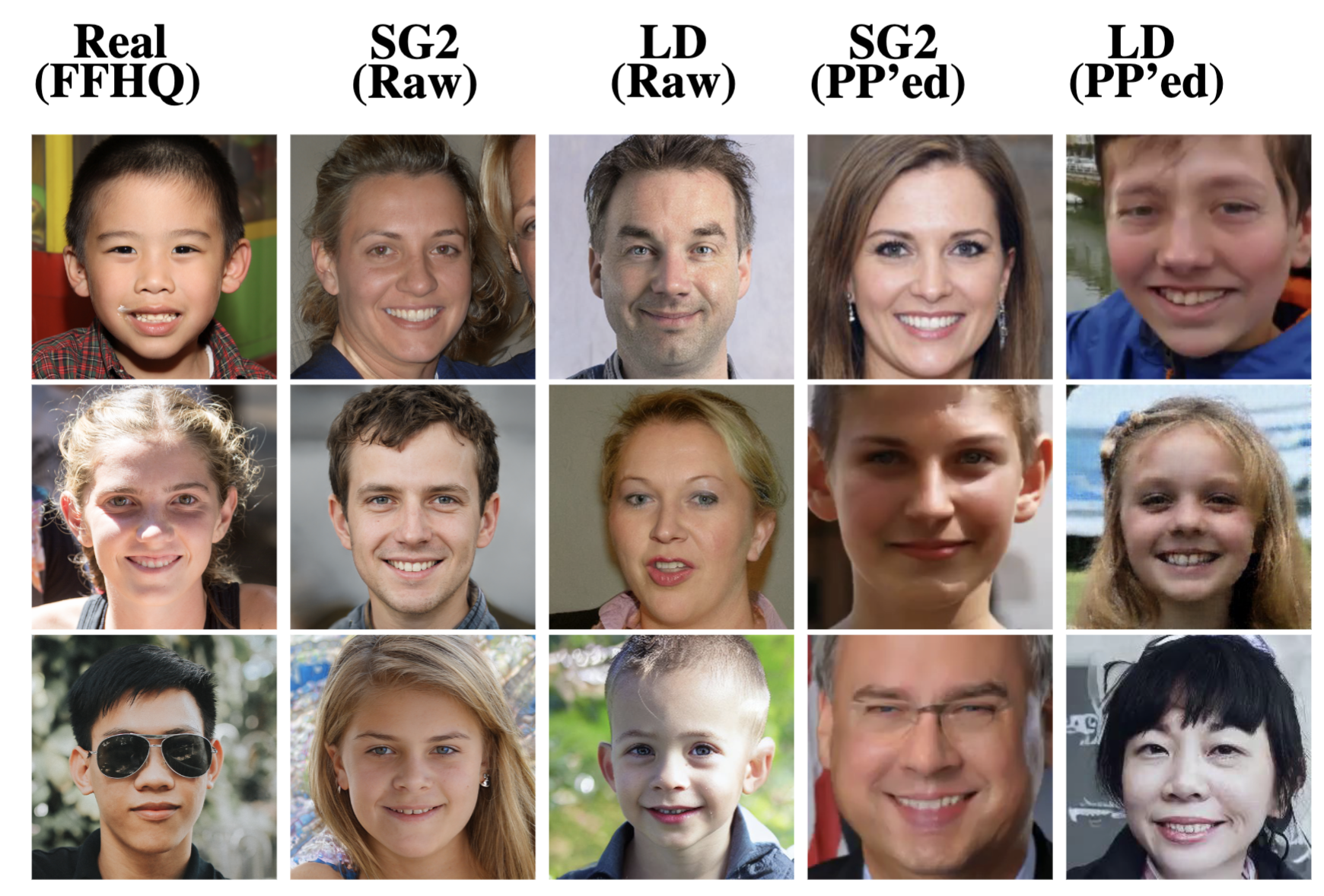}
    \vspace{-0.76cm}
    \caption{\emph{Examples of evaluation data. `SG2' stands for the StyleGAN2 model, `LD' represents the Latent Diffusion model, and `PP'ed' means post-processed data.}}
\vspace{-0.21cm}
\label{fig:data-example}
\end{figure}

\begin{table}[t]
\center
\footnotesize
\caption{\em Detailed information of the evaluation dataset from $DF^3$~\cite{ju2023glff}. `SG2' stands for the StyleGAN2 model, `LD' represents the Latent Diffusion model, and `PP'ed' means post-processed data.}
\vspace{-0.21cm}
\newcommand{\tabincell}[2]{\begin{tabular}{@{}#1@{}}#2\end{tabular}} 
\begin{tabular}{l|c|c|c|c|c}
\hline
  &  \multirow{2}{*}{Real}    & \multicolumn{2}{c|}{Raw} & \multicolumn{2}{c}{PP'ed}     \\ \cline{3-6} 
  &     & SG2 & LD & SG2 & LD \\ \hline
  
Number     & 1,000          & 1,000          & 1,000                    & 1,000            & 1,000                          \\ \hline
Image Size & $512^2$ & $512^2$  &$512^2$  & $256^2$   & $256^2$             \\ \hline
Format & PNG & PNG & JPEG &  PNG, JPEG & PNG, JPEG               \\ \hline
\end{tabular}
\vspace{-0.3cm}
\label{tab:data} 
\end{table}
\smallskip
\noindent\textbf{Data}: Our experiments are based on a set of $1,000$ real face images from the FFHQ dataset \cite{karras2019style} dataset and another $2,000$ images created with generative AI models from the $DF^3$ dataset \cite{ju2023glff}. All images contain a single human face. Two AI generative models are considered, namely StyleGAN2 \cite{karras2020analyzing} and Latent Diffusion \cite{rombach2022high}. We also adopt two evaluation protocols from the $DF^3$ dataset \cite{ju2023glff}. This includes assessing the basic detection performance of raw data and evaluating the robustness of post-processed DeepFake data through mixed operations such as JPEG Compression, Gaussian Blur, face blending, adversarial attacks, and multi-image compression. Detailed information on the data used is given in Table \ref{tab:data}. A few examples of the real and AI-generated images are shown in Fig. \ref{fig:data-example}.

\smallskip
\noindent\textbf{Text Prompts}: Text prompts embody the instruction and request to the LLM to detect DeepFake faces. Properly designed prompts can bring forth the power of semantic knowledge in the LLMs to this task.  We consider prompts of different levels of richness of contexts and additional information in our experiments: 
\begin{itemize}
    \item {\bf Prompt \#1}: {{Tell me if this is an AI-generated image. Answer yes or no.}}
    \item {\bf Prompt \#2}: {{Tell me if this is a real image. Answer yes or no.}} 
    \item {\bf Prompt \#3}: {Tell me the probability of this image being AI-generated. Answer a probability score between 0 and 100.}
    \item {\bf Prompt \#4}: {{Tell me the probability of this image being real. Answer a probability score between 0 and 100.}}
    \item {\bf Prompt \#5}: {Tell me if this is a real or AI-generated image. }
    \item {\bf Prompt \#6}: {Tell me if there are synthesis artifacts in the face or not. Must return with 1) yes or no only; 2) if yes, explain where the artifacts exist by answering in [region, artifacts] form.} 
    \item {\bf Prompt \#7}: {I want you to work as an image forensic expert for AI-generated faces. Check if the image has the artifact attribute listed in the following list and ONLY return the attribute number in this image. The artifact list is [1-asymmetric eye iris; 2-irregular glasses shape or reflection; 3-irregular teeth shape or texture; 4-irregular ears or earrings; 5-strange hair texture; 6-inconsistent skin texture; 7-inconsistent lighting and shading; 8-strange background; 9-weird hands; 10-unnatural edges].}
\end{itemize}
The first two simple binary prompts ask for straightforward Yes/No answers. The third and fourth ones go beyond binary answers and also ask for a numerical value of likelihood. The fifth one makes the LLM to choose between two alternatives of the image to be real or DeepFake. These prompts are {\em simple prompts} and would be a user's first attempt at interacting with LLMs for this task. However, our experiments (detailed later) show that such simple prompts are not effective -- in many cases, the LLM declines to respond to the requests due to a lack of context or safety concerns. When the LLM did respond, the responses were not informative. Prompt \#6 goes beyond simple binary answers: we ask the LLM to identify signs of synthesis and, in addition, request it to justify the answers. This additional request can lead the LLM to be more guided, resulting in the lowest rejection rate. Prompt \#7 goes even further, which includes a more detailed list of clues about possible aspects of which DeepFake faces exhibit semantic inconsistencies. Overall, the more context-rich prompts have lower rates of rejections. On the other hand, the more detailed prompts may lead to lower accuracies. This is possibly because it limits the cues for the LLM to consider, so the LLM may not be able to correctly identify DeepFakes with artifacts not exactly included in the list. In addition, Prompt \#7 uses more tokens (72) than \#6 (31), which increases the cost of running the LLMs. Because of these reasons, we subsequently conducted our experiment based on prompt \#6. 

\begin{figure*}[t]
  \centering
    \includegraphics[width=1.0\textwidth]{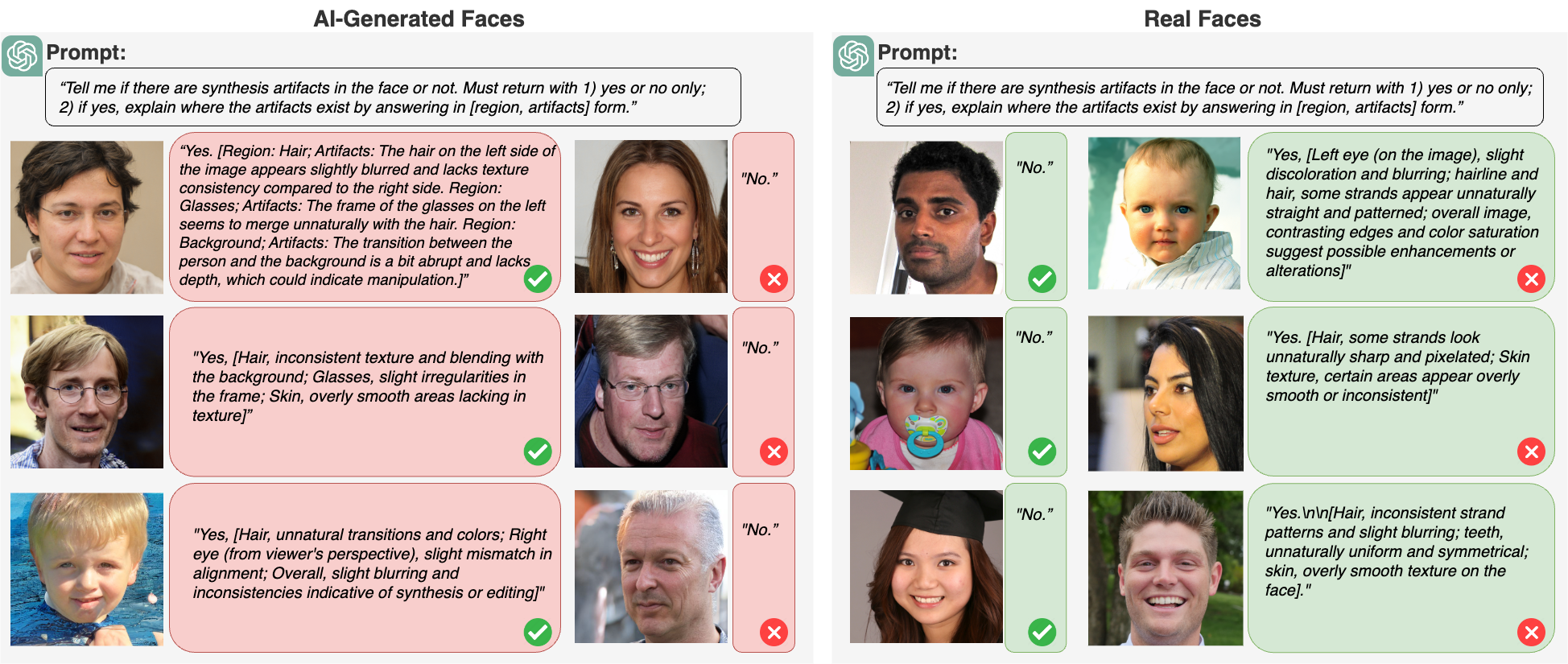}
    \vspace{-0.7cm}
  \caption{\em Examples of \textbf{GPT4V} for DeepFake face detection. Left: Results for AI-generated images from the $DF^3$ dataset~\cite{ju2023glff}. Right: Results for real faces from the FFHQ dataset~\cite{karras2019style}. The responses for AI-generated faces are labeled in \colorbox{color2}{pink}, while those for the real faces are labeled in \colorbox{color3}{green}. Both success (w/ \includegraphics[height=8pt]{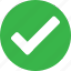}) and failure (w/ \includegraphics[height=8pt]{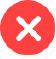}) cases are shown.}
    \vspace{-0.234cm}
  \label{fig:whole}
\end{figure*}

\begin{figure*}[t]
  \centering
    \includegraphics[width=1.0\textwidth]{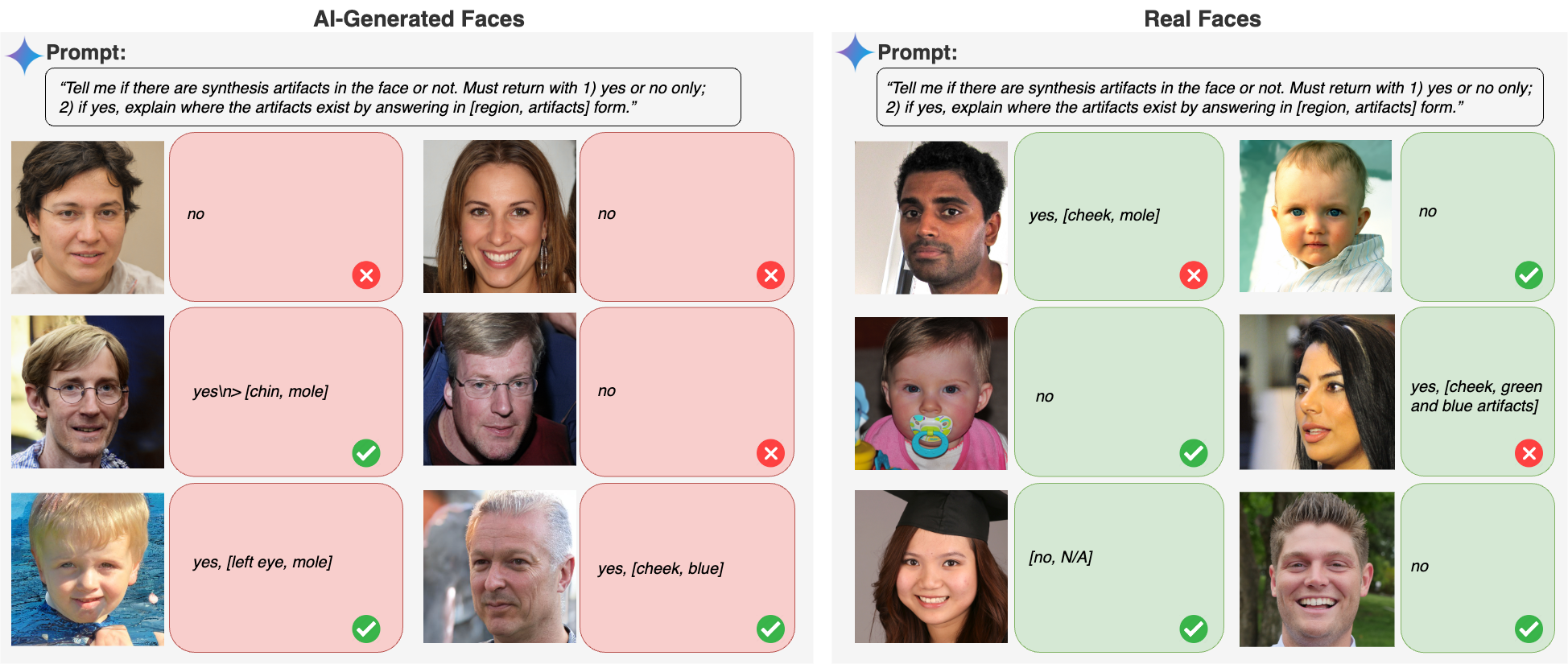}
    \vspace{-0.7cm}
  \caption{\em Examples of \textbf{Gemini 1.0 Pro} for DeepFake face detection. Left: Results for AI-generated images from the $DF^3$ dataset~\cite{ju2023glff}. Right: Results for real faces from the FFHQ dataset~\cite{karras2019style}. The responses for AI-generated faces are labeled in \colorbox{color2}{pink}, while those for real faces are labeled in \colorbox{color3}{green}. Both success (w/ \includegraphics[height=8pt]{figs/correct.pdf}) and failure (w/ \includegraphics[height=8pt]{figs/wrong.pdf}) are shown. We can see that even though some yes/no results are accurate, the supporting evidence is not.}
    \vspace{-0.4cm}
  \label{fig:whole2}
\end{figure*}

\noindent\textbf{Performance Metrics}: For each text-image prompt, we query the LLM multiple times and calculate a numerical score by averaging the results (No $=0$ and Yes $=1$). This approach offers two benefits. Firstly, it diminishes the variability in LLM responses to identical queries, attributable to the probabilistic nature of the underlying LLMs. Secondly, using numerical decision scores enables the application of performance metrics beyond mere accuracy, such as the area under the ROC (AUC) score. Compared to classification accuracies, the AUC score is less affected by imbalanced data, provides a more comprehensive performance evaluation, and allows us to compare the LLM's performance with existing programmed detection methods. AUC score is a real number in $[0,1]$, with higher values corresponding to better performance. As the LLM may decline to respond to a query, another important performance metric is the rejection rate, which measures the fraction of queries that the LLM declines. We also report the single-class accuracy at the fixed threshold of 0.5.

\noindent\textbf{Model Parameters}: All batch tests were performed through API calls. In the evaluation with the GPT4V APIs, we adopted settings similar to those described in \cite{deandrestame2024good}. For the Gemini model, we used Gemini-1.0-pro-vision, which is free of charge and supports up to 60 requests per minute. The total cost of this study is approximately \$130, and it took around 30 days.

\section{Experiment Results}
\label{sec:exp}
\subsection{Qualitative and Quantitative Results}

\definecolor{color1}{rgb}{0.31, 0.43, 0.48}
\definecolor{color2}{rgb}{0.945,  0.816, 0.804}
\definecolor{color3}{rgb}{0.851, 0.906, 0.839}

We show several examples of using GPT4V model with Prompt \#6 to determine if an input image contains a DeepFake face in Fig. \ref{fig:whole}. The left column corresponds to cases when the input images are generated with various AI models, and the right column is for the cases of real images. Both success (with check marks) and failure cases (with crosses) are shown. These results indicate that the GPT4V model achieved a reasonable detection accuracy on this task. We also offer comparison outputs of Gemini 1.0 Pro in Fig. \ref{fig:whole2}, which is less reliable in providing accurate insights for image forensics tasks.

\begin{figure}[t]
    \centering
    \includegraphics[width=0.48\textwidth]{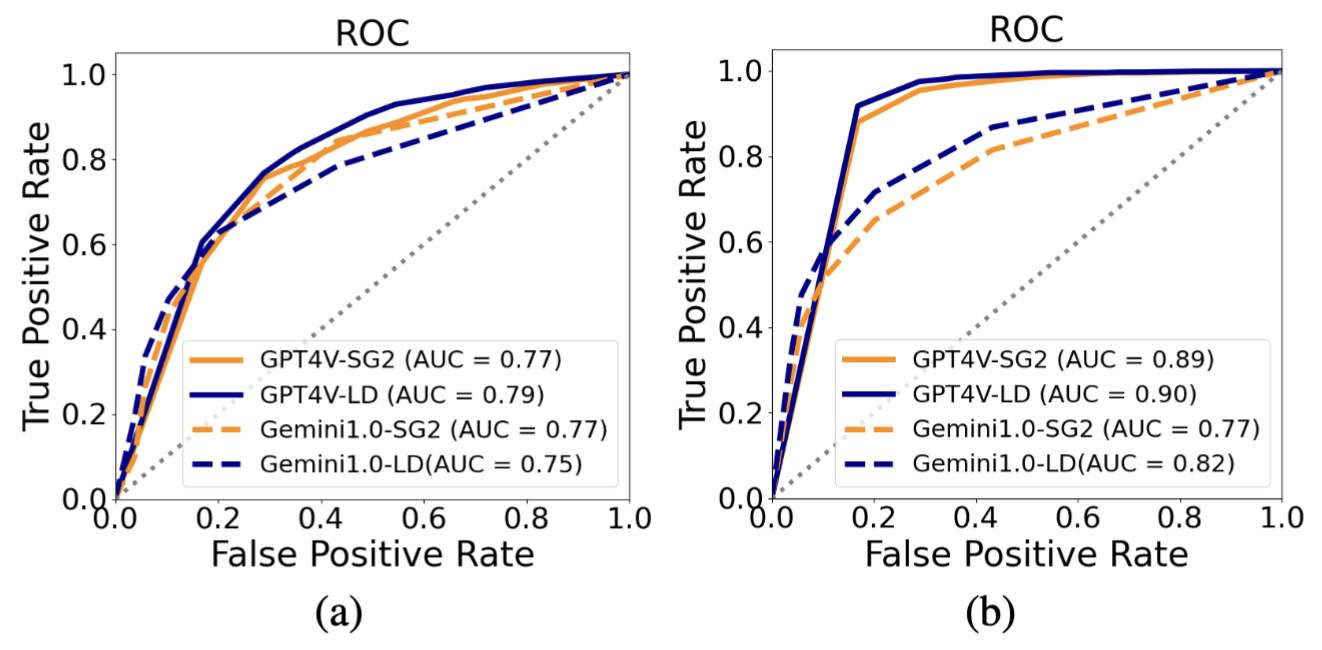}
  \vspace{-0.65cm}
  \caption{\em ROC curves of GPT4V and Gemini 1.0 Pro on the DeepFake detection based on averaging the predictions of five rounds of queries, (a) on raw data, (b) on post-processed DeepFake data.}
  \vspace{-0.421cm}
  \label{fig:roc}
\end{figure}

The quantitative results corroborate this observation. Fig. \ref{fig:roc} shows the {\em receiver operational curves} (ROCs) and the corresponding AUC scores obtained using API calls (as described in Section \ref{sec:method}) over the evaluation dataset with the same prompt -- GPT4V has a $79.5\%$ AUC on raw latent diffusion-generated face images and $77.2\%$ AUC score on StyleGAN-generated face images. The performance confirms that the GPT4V model obviously did not make random guesses on this task (corresponding to a ROC as a diagonal line and a $50\%$ AUC score). Compared to the GPT4V model, Gemini shows a slight decrease in performance.

To put these performances into the context of the state-of-the-art DeepFake face detection methods, we compare them with existing methods in Table \ref{tab:auc} for AUC scores and Table \ref{tab:acc} for classification accuracies. Note that all these baseline detectors were trained on a image forensics dataset \cite{wang2019cnn-generated} with 360K ProGAN-generated images~\cite{karras2017progressive} and 360K real images~\cite{yu2015lsun}. As it shows, the performance of GPT4V and Gemini 1.0 is on par or slightly better than the early methods \cite{wang2019cnn-generated,frank2020leveraging}, but is not competitive with more recent detection methods \cite{gragnaniello2021gan,ju2022fusing,ju2023glff}. This may be attributed to some fundamental aspects between the two approaches. Existing effective DeepFake detection methods can capture signal-level statistical differences between training real and AI-generated images. In contrast, multimodal LLM's decision is mostly based on semantic-level abnormalities, reflected by the additional explanation in natural language in the responses. Therefore, even though the LLM is not specifically designed and trained for DeepFake face detection, the world knowledge encapsulated in the LLM can be transferred to this task. The semantic reasoning leads to results that are more comprehensible to humans. The detection is less susceptible to post-processing operations that can disrupt signal-level features -- this is confirmed with the changes in performances when post-processing is included in Tables \ref{tab:auc} and \ref{tab:acc}, where classification accuracies on DeepFake face even increase for post-processed images. Another factor contributing to this performance enhancement is the inclusion of post-processing operations such as face blending and adversarial attacks, which introduce more distinctive visual artifacts to the images.

\begin{table}[t]
\centering
\caption{\em Comparison of AUC (\%) in detecting DeepFake faces. `SG2' stands for the StyleGAN2 model, and `LD' represents the Latent Diffusion model.}
\vspace{-0.1cm}
\scalebox{0.93}{
\begin{tabular}{l|cc|cc}
\hline
\multirow{2}{*}{Method}        & \multicolumn{2}{c|}{Raw data} & \multicolumn{2}{c}{Post-processed}     \\ \cline{2-5} 
      & SG2 & LD & SG2 & LD \\ \hline
CNN-aug \cite{wang2019cnn-generated}      &   96.5   &  58.6 &  53.2  & 52.4  \\ \hline
GAN-DCT \cite{frank2020leveraging}     & 53.4 &    75.4    &  44.4 &  56.0  \\ \hline
Nodown \cite{gragnaniello2021gan}      & \textbf{99.6}      & \textbf{97.1}   & 47.4     & 44.9   \\ \hline
BeyondtheSpectrum \cite{yang_ijcai21} & 98.1 & 77.3  & 45.4  &  46.9    \\ \hline
PSM \cite{ju2022fusing}  & {99.2} &  82.5  & 73.1  &  71.3   \\ \hline
GLFF \cite{ju2023glff}   & 97.5     & 86.7   & 80.6      & 79.4   \\ \hline \hline
Gemini 1.0 (zero-shot) & 76.6    & 75.1    & 77.5   &   81.5  \\ \hline
GPT4V (zero-shot) & 77.2      &   79.5     & \textbf{88.7}  &  \textbf{89.8}    \\ \hline
\end{tabular}}
\label{tab:auc} 
\end{table}

\begin{table}[]
\centering
\caption{\em Comparison of single-class Accuracy (\%) in detecting DeepFake faces. `SG2' stands for the StyleGAN2 model, and `LD' represents the Latent Diffusion model.}
\vspace{-0.1cm}
\scalebox{0.85}{
\begin{tabular}{l|c|cc|cc}
\hline
\multirow{2}{*}{Method}  & \multirow{2}{*}{Real}       & \multicolumn{2}{c|}{Raw data} & \multicolumn{2}{c}{Post-processed}     \\ \cline{3-6} 
  &    & SG2 & LD & SG2 & LD \\ \hline
CNN-aug \cite{wang2019cnn-generated}   &89.8&	71.9&	0.3&	38.3&	5.5 \\ \hline
GAN-DCT \cite{frank2020leveraging}  & \textbf{92.5} &	3.7  &	7.0&	20.8  &	29.4 \\ \hline
Nodown \cite{gragnaniello2021gan}   &  81.3  &	\textbf{96.3}  &	0.1  &	3.3  &	4.50 \\ \hline
BeyondtheSpectrum \cite{yang_ijcai21} & 67.6  &	42.0 &	8.0 &	11.9  &	15.1 \\ \hline
PSM \cite{ju2022fusing} & 78.0 &	89.8  &	0.1  &	4.4  &	3.3    \\ \hline
GLFF \cite{ju2023glff} & 89.9  &	82.9  &	0.2  &	7.6  &	8.1  \\ \hline \hline
Gemini 1.0 (zero-shot) & 83.3      &   45.1     & 48.2   &  53.2    &  61.2 \\ \hline
GPT4V (zero-shot) &  51.2  &   86.5   &  \textbf{90.3}     & \textbf{98.3}   &   \textbf{99.2}    \\ \hline
\end{tabular}}
  \vspace{-0.21cm}
\label{tab:acc} 
\end{table}


\begin{table*}[]
\centering
\caption{\em Comparison results (\%) of using different prompts for GPT4V in detecting 1,000 StyleGAN2 faces. Note that the Accuracy is measured by comparing the number of correct predictions to the total number of samples that were not rejected.}
\vspace{-0.1cm}
\scalebox{0.92}{
\begin{tabular}{c|c|c|c|c|c|c|c}
\hline
{Metric}         & {Prompt \#1} & {Prompt \#2} & {Prompt \#3} & {Prompt \#4} & {Prompt \#5} & {Prompt \#6} & {Prompt \#7 } \\ \hline
{Rejection Rate} & 60.2             &       66.9         &       100         &       100        &      95.8          &    4.7            &       33.1          \\ \hline
{Accuracy}       &    97.49              &       94.86         &        -       &     -               &       88.10         &     83.83             & 86.54               \\ \hline 
\end{tabular}
\label{tab:promptcom} }
\end{table*}

On the other hand, we note that most errors of GPT4V occur on detecting real images -- per Table \ref{tab:acc}, the classification accuracies on real images are around $50\%$, drastically different from those of AI-generated images, which is above $90\%$. Some intuitions can be obtained when we examine the real face images for these error cases, as shown in Fig. \ref{fig:whole}. These cases include semantic features unusual for ``typical'' face images. For instance, different age group (baby in the first case) or unique hair color (second case) to style (third case). This suggests that the semantic abnormality identified by GPT4V may not be specific to DeepFake faces. This problem may be solved by refining the model. In contrast, the Gemini model achieves a classification accuracy of $83.3\%$ on real images, dropping to around $50\%$ on generated faces. The examples in Fig. \ref{fig:whole2} show that the Gemini model's response lacks rationality in analyzing the synthesis artifacts.

\subsection{Ablation Studies}

\begin{figure*}[t]
\centering
\includegraphics[width=\textwidth]{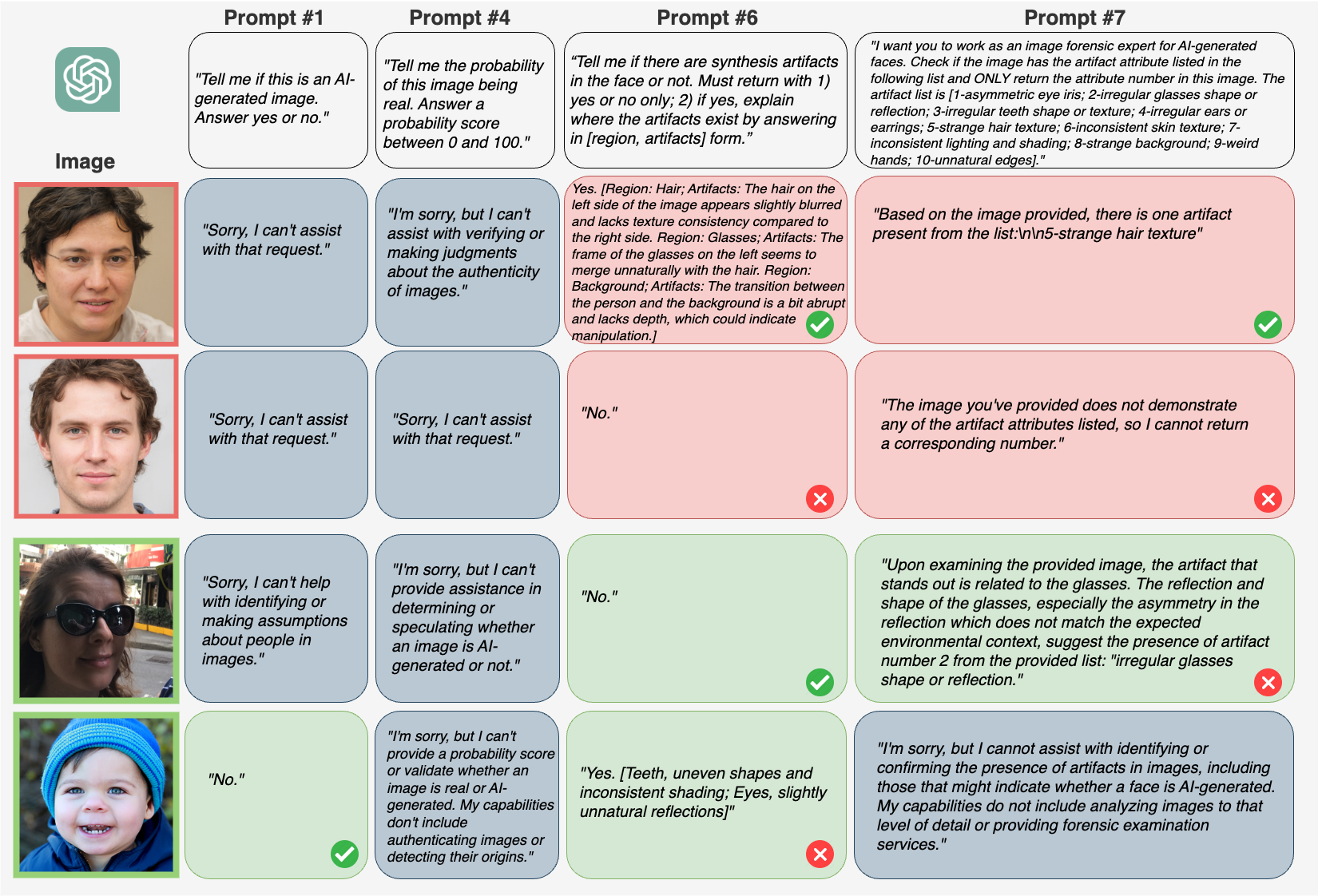}
  \vspace{-0.51cm}
  \caption{\em Examples of {GPT4V} for DeepFake face detection. We show success (w/ \includegraphics[height=8pt]{figs/correct.pdf}) and failure (w/ \includegraphics[height=8pt]{figs/wrong.pdf}), and rejected cases (shown in \colorbox{color1!70}{dark cyan}). The responses for AI-generated faces are labeled in \colorbox{color2}{pink}, while those for the real faces are labeled in \colorbox{color3}{green}. The Figure is best viewed in color. Zoom in and refer to texts for details.}
  \label{fig:prompt}
\end{figure*}
\begin{figure}[t]
\centering
\vspace{-0.34cm}
\includegraphics[width=0.45\textwidth]{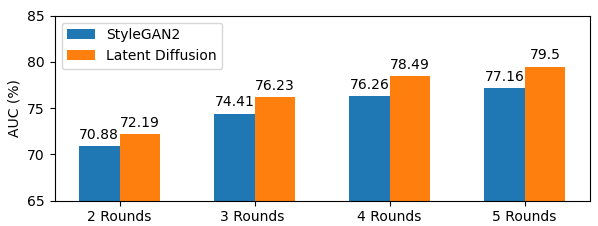}
  \caption{\em Comparative analysis of AUC scores (\%) across different query rounds of GPT4V in DeepFake Detection.}
\vspace{-0.44cm}
  \label{fig:round}
\end{figure}

\begin{figure}[t]
\centering
\vspace{-0.3cm}
\includegraphics[width=0.482\textwidth]{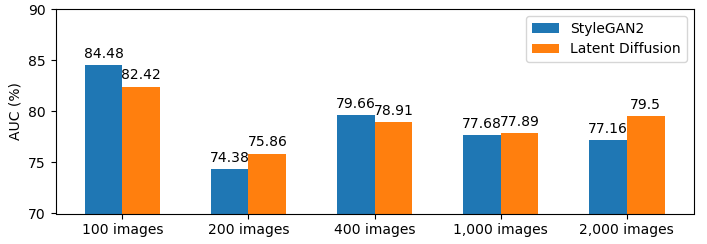}
  \caption{\em Comparative analysis of AUC scores (\%) using different data size of GPT4V in DeepFake Detection.}
\vspace{-0.4cm}
  \label{fig:size}
\end{figure}
\begin{figure*}[t]
\centering
\includegraphics[width=\textwidth]{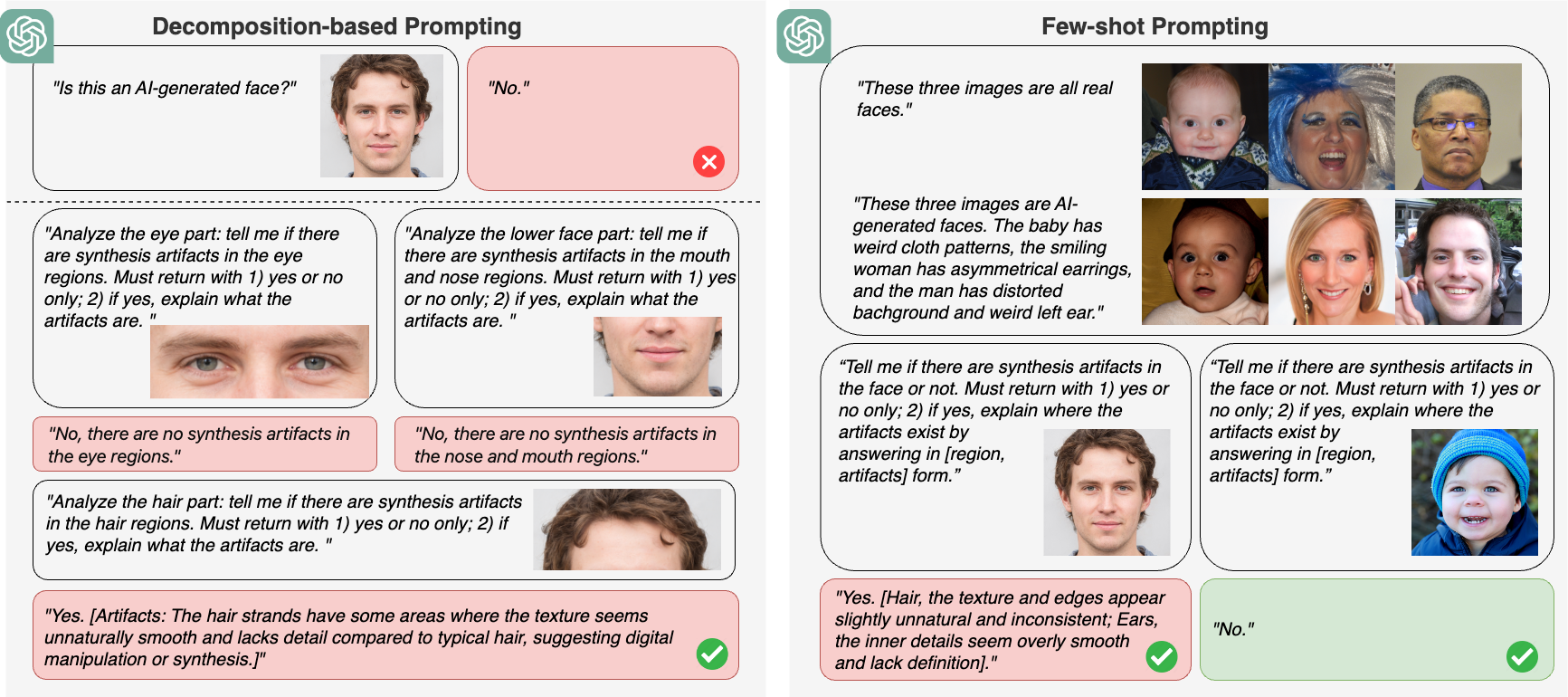}
\vspace{-1.3em}
  \caption{\em Potential improvement in detecting DeepFake images. The responses for AI-generated faces are labeled in \colorbox{color2}{pink}, while those for the real faces are labeled in \colorbox{color3}{green}. Success case (w/ \includegraphics[height=8pt]{figs/correct.pdf}) and failure case (w/ \includegraphics[height=8pt]{figs/wrong.pdf}) are shown.}
  ~\vspace{-1em}
  \label{fig:prompt_impro}
\end{figure*}
The quality of the prompt plays a central role in performance. In addition to the prompts used in the experiments, we have also studied other prompts with simpler structures and compared their performance.
Firstly, we quantitatively compare different text prompts in detecting 1,000 raw StyelGAN2 faces. Table \ref{tab:promptcom} reports the rejection rate and accuracy of GPT4V with all seven prompts described in Section \ref{sec:method}. The findings indicate that prompts related to direct image forensics result in high rejection rates, particularly those based on likelihood assessments and prompts requiring a choice between real or fake. Prompts \#6 and \#7 result in fewer rejections with comparable prediction accuracies because they extend beyond mere yes-or-no responses by asking the model to identify signs of synthesis. Fig. \ref{fig:prompt} shows four examples predicted by GPT4V using different prompts. GPT4V misclassified visually realistic fake faces and interprets unusual semantic features in real faces as synthesis artifacts. Next, we show the influence of the number of query attempts on detection performance. Fig. \ref{fig:round} demonstrates that increasing query attempts correlates with higher AUC scores. This indicates that repeated querying might serve as an ensemble method for enhancing performance.
Finally, we explore how the dataset size affects the detection performance of GPT4V. Fig. \ref{fig:size} presents the comparison results using different numbers of evaluation data, with each set containing an equal balance of real and generated images. As the dataset grows, the performance for the StyleGAN2 and Latent Diffusion models tends to converge.
\vspace{-0.12cm}
\subsection{Improvements}
\vspace{-0.1cm}
So far, we have only tested simple queries. It has been demonstrated that using better prompts constructed with {chain-of-thought prompts} \cite{wei2022chain}, few-shot prompting~\cite{yang2023dawn}, which provide step-to-step guides in an interactive conversation with the LLM can elicit more relevant responses.
The API interface of GPT4V and Gemini 1.0 does not allow multiple rounds of dialog because no consistent context is stored across API calls. Therefore, such interactive guidance can only be used with the web interface through a manual interaction, \ie, they cannot be automated using API calls. We provide two exploratory approaches in Fig. \ref{fig:prompt_impro}: one employs decomposed local images, a method we refer to as decomposition-based prompting, and the other utilizes a few-shot prompting technique. By supplying decomposed parts of images, we direct the model's attention towards finer local patterns and reveal subtle visual anomalies. Further, using few-shot prompting with brief synthesis instructions imparts crucial forensic knowledge to the model, enabling the model to correctly classify two samples that were previously misidentified. 
These initial results suggest that using more crafted prompts can improve performance. However, we will wait for the LLMs to enable consistent API calls.

\section{Conclusion}
\vspace{-0.11cm}
\label{sec:con}

In this study, we investigate the potential of leveraging multimodal LLMs for tasks related to media forensics. Our future research will broaden the application of multimodal LLMs to include a wider array of media forms, particularly focusing on video analysis. Rather than simply applying image-based detection techniques to video frames, a more integrated approach would involve direct video content processing. Additionally, we aim to enhance the detection of text-image mis-contextualization \cite{huang2023exposing}, where images and text are misleadingly paired to spread false information. Future endeavors will also explore developing more sophisticated prompting strategies and integrating these models with conventional signal or data-driven detection techniques.

\noindent{\bf Impact Statement}. This work explores the use of multimodal LLMs in media forensics. We realize that LLMs may hallucinate information due to the biases in training data. Therefore, human users should always verify the results to avoid potential mistakes.

\noindent{\bf Acknowledgement}. Siwei Lyu is supported by U.S. National Science Foundation (NSF) under grant SaTC-2153112. Xin Li is supported by U.S. NSF under grant CCSS-2348046 and SUNY-Albany start-up funds. Baoyuan Wu is supported by National Natural Science Foundation of China under grant No.62076213, Shenzhen Science and Technology Program under grant No.RCYX20210609103057050, and the Longgang District Key Laboratory of Intelligent Digital Economy Security. 
{
    \small
    \bibliographystyle{ieeenat_fullname}
    \bibliography{main, misc}
}


\end{document}